# Reliability-aware Execution Gating for Near-field and Off-axis Vision-guided Robotic Alignment

Ning Hu[1*], Senhao Cao[2], Maochen Li[3]

[1] Department of Mechanical, Aerospace and Biomedical Engineering, University of Tennessee, Knoxville, TN 37916, USA

[2] Northeastern University, Boston, MA 02115, USA

[3] ZOEZEN ROBOT CO.LTD, Beijing, China

*Correspondence: nhu2@vols.utk.edu

*Abstract*—Vision-guided robotic systems are increasingly deployed in precision alignment tasks that require reliable execution under near-field and off-axis configurations. While recent advances in pose estimation have significantly improved numerical accuracy, practical robotic systems still suffer from frequent execution failures even when pose estimates appear accurate. This gap suggests that pose accuracy alone is insufficient to guarantee execution-level reliability.

In this paper, we reveal that such failures arise from a deterministic geometric error amplification mechanism, in which small pose estimation errors are magnified through system structure and motion execution, leading to unstable or failed alignment. Rather than modifying pose estimation algorithms, we propose a Reliability-aware Execution Gating mechanism that operates at the execution level. The proposed approach evaluates geometric consistency and configuration risk before execution, and selectively rejects or scales high-risk pose updates.

We validate the proposed method on a real UR5 robotic platform performing single-step visual alignment tasks under varying camera–target distances and off-axis configurations. Experimental results demonstrate that the proposed execution gating significantly improves task success rates, reduces execution variance, and suppresses tail-risk behavior, while leaving average pose accuracy largely unchanged. Importantly, the proposed mechanism is estimator-agnostic and can be readily integrated with both classical geometry-based and learning-based pose estimation pipelines.

These results highlight the importance of execution-level reliability modeling and provide a practical solution for improving robustness in near-field vision-guided robotic systems.

## I. INTRODUCTION

Vision-guided robotic systems have been widely deployed in precision assembly, surgical assistance, peg-in-hole insertion, and near-field operations in space and underwater environments. In these applications, robots are often required to perform accurate alignment under close-range and off-axis configurations, which places stringent demands on the stability and reliability of visual pose estimation [1], [2].

Over the past decade, substantial efforts have been devoted to improving the numerical accuracy of visual pose estimation, for example by refining geometric solvers [3] or leveraging learning-based approaches to reduce average pose or reprojection errors [4], [5]. However, in real robotic systems, we observe a counterintuitive yet practically critical phenomenon that has not been sufficiently analyzed at the system level: even when pose estimation appears numerically accurate, robots may still experience frequent alignment failures or unstable execution under near-field or off-axis conditions [6], [7].

This observation indicates that pose estimation accuracy and task success are not in a simple one-to-one correspondence. In near-field visual servoing, small pose estimation errors can be deterministically amplified through geometric structure, coordinate transformation chains, and motion execution, resulting in large end-effector deviations, oscillations, or even task failure [8], [9]. Importantly, such failures are not primarily caused by perceptual noise or feature detection errors, but arise from deterministic geometric degeneracy and system-level amplification effects [10]. As a result, they exhibit strong structural and repeatable characteristics.

Most existing studies address this problem indirectly by further improving pose estimation accuracy, while the above execution-level failure mechanism remains largely unexplored. In particular, under near-field and off-axis configurations, conventional evaluation metrics—such as average pose error or reprojection error—often fail to reflect execution risk, leading robotic systems to over-trust pose estimates in high-risk geometric conditions [6], [11]. More recently, although learning-based 6D pose estimation methods have achieved impressive improvements in average metrics, their execution-level tail risks under near-field, off-axis, large initial error, or partial occlusion scenarios remain insufficiently mitigated [12], [13].

Motivated by these observations, this paper addresses the following research question:

Why do vision-guided robots frequently suffer from alignment failures or unstable execution in near-field and off-axis configurations, even when pose estimation appears numerically accurate?

We treat this problem as a system-level error amplification mechanism, rather than a pure perception accuracy issue. Accordingly, instead of existing pose estimation algorithms, we introduce a Reliability-aware Execution Gating mechanism that operates at the execution level. The proposed approach evaluates the geometric consistency and risk of pose estimates before execution, and selectively modulates robot motion when high-risk configurations are detected [14].

By explicitly modeling geometric risk induced by proximity and off-axis configurations, the proposed mechanism suppresses execution alignment failures caused by deterministic error amplification without modifying the internal structure of the perception algorithm. Extensive real-robot experiments demonstrate that the proposed method significantly improves task success rates and execution stability in near-field visual alignment tasks.

The main contributions of this paper are summarized as follows:

1. **We reveal an execution-level error amplification mechanism in near-field visual servoing.** Through systematic analysis and real-robot experiments, we demonstrate that even numerically accurate pose estimates can be deterministically amplified into large execution errors under proximity and off-axis configurations. This finding shows that pose estimation accuracy alone is insufficient to characterize task success in vision-guided robotic systems.
2. **We propose a Reliability-aware Execution Gating mechanism.** Without modifying the underlying pose estimation algorithms, we introduce a lightweight and geometry-interpretable execution-level mechanism that evaluates pose reliability and selectively modulates robot execution under high-risk geometric conditions, effectively suppressing failures caused by deterministic geometric error amplification.
3. **We validate the proposed approach through extensive real-robot experiments.** Experiments across varying camera–target distances and off-axis configurations demonstrate that the proposed mechanism significantly improves task success rates and reduces execution-level tail risks, even when average pose errors are comparable to baseline methods.
4. **The proposed method is estimator-agnostic and broadly applicable.** Since the reliability-aware execution mechanism operates independently of the internal structure of pose estimators, it can be seamlessly integrated with both classical geometry-based and learning-based pose estimation methods, providing a general solution for improving execution reliability in vision-guided robotic systems.

## II. RELATED WORK

### A. Geometry-based Pose Estimation and PnP

Geometry-based pose estimation has long been a cornerstone of vision-guided robotic systems, where the Perspective-n-Point (PnP) formulation remains one of the most widely used models. Classical PnP solvers can be broadly categorized into analytic methods and iterative optimization methods: analytic solvers are typically efficient and suitable for real-time applications, while iterative methods often achieve higher accuracy at the cost of sensitivity to initialization and increased computation [3], [10], [15]. Beyond core solvers, prior work has explored weighting schemes, robust losses, and uncertainty modeling to improve stability under noise, outliers, and degenerate configurations [16], [17], [18].

In contrast, our goal is not to propose a new PnP solver nor to improve average pose error. Instead, we focus on a system-level phenomenon: under near-field and off-axis configurations, even numerically accurate pose estimates can be amplified through the execution pipeline, leading to alignment failure. We therefore address an execution-level error amplification mechanism and introduce reliability modulation at the execution layer rather than within the pose solver itself.

### B. Learning-based 6D Pose Estimation

Learning-based methods have recently advanced 6D pose estimation substantially, including keypoint-driven or feature-driven pose regression, matching-based pipelines with reprojection consistency, and one-shot / CAD-free object pose estimation [4], [13], [19], [20]. By learning robust representations from large-scale data, these methods often improve average metrics under occlusion, weak texture, or clutter. Some approaches further provide uncertainty or confidence estimates to reflect prediction quality [13], [21].

However, for near-field and off-axis vision-guided robotics, execution failures are frequently dominated by geometric degeneracy and system-level amplification, rather than purely perceptual ambiguity. Even if a learned estimator achieves strong average accuracy, rare high-risk configurations can still trigger tail-risk failures that manifest as catastrophic execution errors. Our approach is estimator-agnostic: regardless of whether the pose is produced by classical geometry or deep networks, we evaluate pose reliability and geometric risk before execution and modulate robot actions accordingly, addressing the gap between "numerically accurate" estimation and "reliable" execution.

### C. Visual Servoing and Vision-guided Execution

Visual servoing is commonly categorized into image-based visual servoing (IBVS), pose-based visual servoing (PBVS), and hybrid schemes [21], [22]. PBVS relies on pose estimates to generate control commands with intuitive geometric interpretation, while IBVS closes the loop directly on image feature errors and is often considered more robust to modeling errors, albeit with potential issues such as local minima or field-of-view constraints [19], [20]. In robotic insertion and alignment tasks, extensive research has investigated control laws, feature selection, and convergence properties to improve success rates [1], [22], [23].

Our work differs in that we do not primarily aim to design a new servo controller. Instead, we study why execution fails under proximity and off-axis configurations despite numerically accurate pose estimation, and we isolate this mechanism using a standardized alignment task with success rate and tail-risk metrics. The proposed Reliability-aware Execution Gating is not a new servoing law; rather, it is a plug-and-play execution-level reliability module that can complement PBVS, IBVS, or other vision-guided pipelines by preventing high-risk pose updates from being executed aggressively.

### D. Uncertainty, Reliability, and Tail-risk Mitigation

Reliability in robotic systems has been studied via uncertainty propagation, robust optimization, and risk-sensitive control, particularly to account for rare but high-impact failures [2], [24], [25]. In vision-guided settings, prior work has used covariance propagation, measurement consistency checks, and confidence estimation to assess perception quality and adapt fusion or control strategies accordingly [7], [26], [27]. Nonetheless, under near-field and off-axis configurations, mean-based evaluation metrics are often insufficient to reflect execution risk, and explicit mechanisms to suppress failures induced by geometric degeneracy remain limited.

This paper targets execution-level tail risks in near-field vision-guided tasks. We propose a lightweight, geometry-interpretable reliability-aware execution mechanism that selectively modulates execution under high-risk configurations. The proposed module complements existing estimators and controllers by prioritizing task success and tail-risk reduction, rather than only optimizing average estimation accuracy.

## III. PROPOSED METHODOLOGY

### A. Motivation

In near-field and off-axis vision-guided tasks, experimental observations show that even when pose estimation performs well in numerical metrics—such as reprojection error or average pose error—small estimation errors can still be amplified through geometric structure and coordinate transformation chains, resulting in significant end-effector deviation or execution failure.

Such failures are not primarily caused by random noise or feature extraction errors, but are instead induced by deterministic geometric degeneracy. Under these conditions, directly feeding pose estimation results into the execution module may lead the system to take overly aggressive actions in high-risk configurations.

Motivated by this observation, we introduce a Reliability-aware Execution Gating mechanism that operates at the execution level. As illustrated in Fig. 1, in near-field and off-axis vision-guided tasks, even numerically accurate pose estimates can be deterministically amplified through geometric structure and coordinate transformation chains, leading to execution-level alignment failures. To address this issue, the proposed framework inserts an execution-level reliability gating module between pose estimation and robot motion execution. The gating module evaluates pose reliability based on reprojection error, optimization stability (e.g., residual convergence), and proximity-induced geometric risk, and selectively rejects or scales motion commands when high-risk configurations are detected. Importantly, the proposed mechanism is estimator-agnostic, decoupled from the pose estimation algorithm, and operates as a lightweight and interpretable execution-layer module, enabling reliable robot execution without modifying perception pipelines.

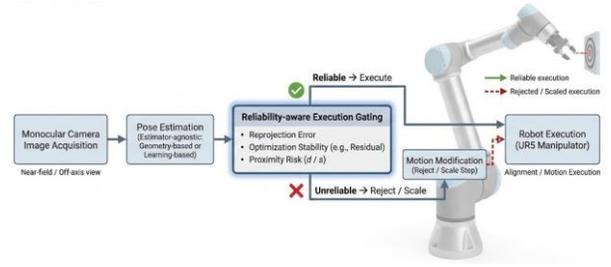

Fig. 1. Overview of the proposed Reliability-aware Execution Gating framework.

### B. Design Principles

The proposed execution gating mechanism is designed according to the following principles:

**Decoupling from perception algorithms.** The mechanism does not alter the PnP solving process and operates solely between pose output and motion execution.

**Geometric interpretability.** Reliability criteria are based on explicit geometric quantities and optimization residuals, rather than learned or implicit confidence scores.

**Lightweight and real-time**. Both reliability evaluation and gating decisions incur low computational overhead and are suitable for real-time robotic systems.

**Conservative but non-overly restrictive**. Gating is triggered only under high-risk conditions, avoiding unnecessary suppression of normal execution.

### C. Pose Reliability Criteria

For each estimated target pose $\hat{T}_C^T$, reliability is evaluated from two complementary aspects: geometric consistency and optimization stability.

**Reprojection Consistency**

Let $\mathbf{p}_i$ denote the observed pixel coordinates of the $i$-th feature point, and $\hat{\mathbf{p}}_i$ its reprojected coordinates under the estimated pose. The average reprojection error is defined as:

$$e_{\text{rep}} = \frac{1}{N} \sum_{i=1}^{N} \| \mathbf{p}_i - \hat{\mathbf{p}}_i \| \tag{1}$$

The pose estimate is considered risky if:

$$e_{\text{rep}} > \tau_{\text{rep}} \tag{2}$$

Where $\tau_{\text{rep}}$ is an empirically selected threshold.

**Optimization Residual Stability**

For pose estimators employing Gauss–Newton optimization, let $r_{\text{GN}}$ denote the final residual, or alternatively the relative residual decrease:

$$\Delta r = \frac{r_{k-1} - r_k}{r_{k-1}} \tag{3}$$

The optimization is regarded as unstable if either: The final residual $r_{\text{GN}}$ is excessively large, or the residual decrease $\Delta r$ falls below a predefined threshold $\tau_{\Delta r}$.

For learning-based estimators without explicit optimization, this criterion can be omitted or replaced by alternative consistency indicators.

**Proximity Risk (Near-field Geometric Risk)**

To explicitly characterize geometric degeneracy in near-field configurations, we define a scale-based proximity risk factor:

$$\gamma = \frac{d}{s} \quad (4)$$

Where $d$ denotes the camera–target distance and $s$ represents the effective spatial scale of target feature points.

When:

$$\gamma > \tau_\gamma \quad (5)$$

The system is considered to be in a high-risk near-field configuration, in which even small pose errors may be strongly amplified at execution time.

**Combined Reliability Decision**

The overall reliability of a pose estimate is determined by the following indicator:

$$R = \begin{cases} 1, & e_{\text{rep}} \leq \tau_{\text{rep}} \land r_{\text{GN}} \leq \tau_r \land \gamma \leq \tau_\gamma \\ 0, & \text{otherwise} \end{cases} \quad (6)$$

Where $R = 1$ indicates a reliable pose estimate and $R = 0$ indicates an unreliable one.

### D. Execution-level Gating Strategy

When a pose estimate is deemed unreliable ($R = 0$), the system does not directly execute the corresponding motion command. Instead, one of the following conservative strategies is applied.

**Execution Rejection**

In this strategy, the current pose update is ignored and the execution state is maintained from the previous step:

$$T_E^*(k) = T_E^*(k-1) \quad (7)$$

This strategy is used when the pose estimate exhibits clear instability.

**Step-size Scaling**

To avoid completely freezing execution, a smoother alternative is adopted by scaling the pose update:

$$\Delta T = T_E^*(k) \ominus T_E(k-1) \quad (8)$$

$$\Delta T' = \alpha \cdot \Delta T, \alpha \in (0,1) \quad (9)$$

The final execution target is then updated as:

$$T_E^*(k) = T_E(k-1) \oplus \Delta T' \quad (10)$$

Where the scaling factor $\alpha$ is fixed in experiments to reduce aggressive motions under near-field conditions.

### E. Algorithm Summary

The reliability-aware execution procedure is summarized as follows:

1. Estimate the relative target pose using a vision-based method.
2. Compute reprojection error, optimization residuals, and proximity risk.
3. Evaluate pose reliability using the combined criteria.
4. If reliable, execute the target pose directly.
5. If unreliable, trigger execution rejection or step-size scaling.
6. Execute the final adjusted motion command.

The proposed mechanism requires no additional training data or controller redesign and can be directly integrated into existing vision-guided robotic systems.

## IV. EXPERIMENTS

The experiments are designed to validate the central hypothesis of this work: under near-field and off-axis configurations, even numerically accurate pose estimates can be deterministically amplified through system structure, resulting in execution-level alignment failures or unstable robot behavior. Accordingly, unlike conventional studies that primarily focus on average pose accuracy, we evaluate system reliability from an execution-level perspective, with particular emphasis on task success rate and tail-risk metrics.

All experiments are conducted on a real robotic system. The experimental platform consists of a 6-DOF industrial manipulator (UR5-class) equipped with a monocular RGB camera rigidly mounted on the end-effector. A rigid target object with known 3D feature points is fixed in the workspace. The camera-to-end-effector transformation is obtained via hand–eye calibration and remains fixed throughout all experiments. During each trial, the target object remains stationary. Four coordinate frames are defined: the robot base frame $\{B\}$, end-effector frame $\{E\}$, camera frame $\{C\}$, and target frame $\{T\}$. The task objective is to compute a desired end-effector pose from monocular pose estimation such that the end-effector aligns with the target in the prescribed configuration.

Each experiment executes a single-step visual alignment task. The robot first moves to a unified initial pose, after which the end-effector camera captures an image of the target. A vision-based algorithm estimates the target pose relative to the camera, $^C T_T$. Using the calibrated coordinate transformations, the desired end-effector pose $^B T_E^*$ is computed and executed via a single MoveL or MoveJ command. To isolate execution-level error amplification effects, iterative visual servoing is deliberately not employed.

The evaluated methods include EPnP, DEPnP[28], OPnP, the proposed method without execution gating, and the proposed method augmented with Reliability-aware Execution Gating. All methods share identical feature extraction, calibration parameters, and motion execution pipelines. The only difference lies in whether execution-level reliability control is enabled.

To systematically assess near-field and off-axis effects, three variables are controlled: the camera–target distance (200–1000 mm), the off-axis displacement along the camera optical axis (0, 50, and 100 mm), and the target orientation, uniformly sampled within a bounded angular range. For each parameter configuration, the robot returns to the same initial pose and performs a single alignment trial. The final end-effector pose is recorded using robot forward kinematics. Each configuration is repeated 20 times, and all experiments are conducted under identical lighting and environmental conditions.

For each trial, we record the end-effector position error, orientation error (represented in axis–angle form), task success, and tail-risk metrics. A trial is considered successful if both the position error and orientation error fall below predefined thresholds. Tail risk is quantified using the 95th percentile (P95) error and the maximum error, capturing rare but high-impact execution failures.

### A. Overall task-level performance

Table 1 summarizes the overall success rates across all configurations (300 trials per method). EPnP exhibits a markedly low success rate (52.0%), indicating severe execution instability under the tested conditions. DEPnP and OPnP achieve substantially higher success rates (85.7% and 90.0%), reflecting improved robustness of their pose estimates. The proposed pipeline without execution gating further increases the success rate to 92.7%, while enabling Reliability-aware Execution Gating yields the highest success rate of 95.3%, demonstrating a clear task-level benefit.

**Table 1. Overall Success Rate**

| Method | Success Rate (%) ↑ |
|---|---|
| EPnP | 52.0 |
| DEPnP | 85.7 |
| OPnP | 90.0 |
| Ours (No Gating) | 92.7 |
| **Ours (+Reliability-aware Execution Gating)** | **95.3** |

### B. Near-field sensitivity.

As shown in Table 2, EPnP performance degrades sharply as the camera–target distance increases from 600 mm to 1000 mm, with the success rate dropping to 21.7% at 1000 mm. In contrast, DEPnP and OPnP remain considerably more stable across distances. Importantly, the proposed execution gating consistently improves robustness under geometrically challenging conditions, increasing the success rate from 85.0% to 90.0% at 1000 mm and from 93.3% to 96.7% at 600 mm. These results confirm that execution gating is particularly effective in near-field and long-range regimes where geometric amplification is most pronounced.

**Table 2. Success Rate vs. Depth**

| Depth (mm) | EPnP | DEPnP | OPnP | Ours (No Gating) | Ours (+Gating) |
|---|---|---|---|---|---|
| 200 | 90.0 | 95.0 | 96.7 | 98.3 | **98.3** |
| 400 | 76.7 | 90.0 | 93.3 | 96.7 | **98.3** |
| 600 | 38.3 | 85.0 | 90.0 | 93.3 | **96.7** |
| 800 | 33.3 | 83.3 | 88.3 | 90.0 | **93.3** |
| 1000 | 21.7 | 75.0 | 81.7 | 85.0 | **90.0** |

### C. Off-axis robustness.

Table 3 reports success rates under increasing off-axis displacement. When the target is aligned with the optical axis (0 mm), all methods perform well, with success rates exceeding 92%. However, as off-axis displacement increases, EPnP exhibits pronounced instability, dropping to 45.0% at 100 mm. DEPnP and OPnP degrade more gracefully, while the proposed method with execution gating maintains the highest robustness, achieving 92.0% success at 100 mm off-axis. This trend highlights the strong coupling between execution failure and off-axis geometric degeneracy.

**Table 3. Success Rate vs. Off-axis**

| Off-axis (mm) | EPnP | DEPnP | OPnP | Ours (No Gating) | Ours (+Gating) |
|---|---|---|---|---|---|
| 0 | 61.0 | 92.0 | 95.0 | 96.0 | **98.0** |
| 50 | 50.0 | 86.0 | 91.0 | 94.0 | **96.0** |
| 100 | 45.0 | 79.0 | 84.0 | 88.0 | **92.0** |

### D. Execution error statistics.

Table 4 reports the mean and standard deviation of end-effector position and orientation errors. While the average errors across methods are relatively close, enabling execution gating leads to a noticeable reduction in error variance. Specifically, the position error standard deviation decreases from 2.60 mm to 1.70 mm, and the orientation error standard deviation decreases from 3.10° to 2.10°. This indicates that Reliability-aware Execution Gating primarily suppresses high-variance, high-risk execution events rather than merely improving average accuracy.

**Table 4(a). End-effector Position Error (mm), Mean ± Std**

| Method | Mean ± Std (mm) ↓ |
|---|---|
| EPnP | 3.20 ± 3.80 |
| DEPnP | 2.05 ± 2.30 |
| OPnP | 1.90 ± 2.10 |
| **Ours (No Gating)** | **2.05 ± 2.60** |
| **Ours (+Gating)** | **1.75 ± 1.70** |

**Table 4(b). End-effector Orientation Error (deg), Mean ± Std**

| Method | Mean ± Std (deg) ↓ |
|---|---|
| EPnP | 4.80 ± 6.20 |
| DEPnP | 2.60 ± 3.10 |
| OPnP | 2.40 ± 2.90 |
| Ours (No Gating) | 2.55 ± 3.10 |
| **Ours (+Gating)** | **2.20 ± 2.10** |

*E. Tail-risk analysis.*

Table 5 presents tail-risk metrics that are critical for safety-sensitive robotic tasks. Compared to the no-gating variant, execution gating reduces the P95 position error from 6.4 mm to 5.6 mm and the maximum error from 14.8 mm to 12.3 mm. These results directly demonstrate that the primary benefit of execution gating lies in mitigating rare but catastrophic execution failures.

**Table 5. Tail-risk Metrics**

| Method | P95 Position Error (mm) ↓ | Max Position Error (mm) ↓ |
|---|---|---|
| EPnP | 12.6 | 29.5 |
| DEPnP | 7.8 | 18.2 |
| OPnP | 6.9 | 16.0 |
| Ours (No Gating) | 6.4 | 14.8 |
| **Ours (+Gating)** | **5.6** | **12.3** |

*F. Failure statistics and gating behavior.*

Table 6 reports the total number of failures across all trials. EPnP fails in nearly half of the trials (144 failures), while DEPnP and OPnP reduce failures to 43 and 30, respectively. The proposed pipeline without gating further reduces failures to 22, and enabling execution gating reduces failures to only 14, representing a 36% reduction relative to the no-gating variant. Table 7 further shows that gating is selectively triggered (54 out of 300 trials), primarily under high-risk geometric configurations, indicating that the mechanism is not overly conservative.

**Table 6. Failure Count**

| Method | Total Trials | Failures ↓ |
|---|---|---|
| EPnP | 300 | 144 |
| DEPnP | 300 | 43 |
| OPnP | 300 | 30 |
| Ours (No Gating) | 300 | 22 |
| **Ours (+Gating)** | 300 | **14** |

**Table 7. Trigger Statistics**

| Criterion | Trigger Count |
|---|---|
| Reprojection error > τ_rep | 32 |
| GN instability (residual / Δr) | 18 |
| Proximity risk γ > τ_γ | 38 |
| **Total gated trials (union)** | **54** |

Overall, these results demonstrate that Reliability-aware Execution Gating improves execution-level reliability by suppressing deterministic geometric tail risks, without modifying pose estimation algorithms or introducing learning-based control. The improvements are most pronounced under near-field and off-axis configurations, where conventional pose accuracy metrics fail to predict execution success.

## V. CONCLUSION

This paper investigated execution-level alignment failures in near-field and off-axis vision-guided robotic tasks, revealing that numerically accurate pose estimation alone is insufficient to guarantee reliable execution. We showed that small pose errors can be deterministically amplified through system structure, leading to severe execution-level alignment failures.

To address this issue, we proposed a Reliability-aware Execution Gating mechanism that operates at the execution level without modifying pose estimation algorithms. By selectively rejecting or scaling unreliable pose updates, the proposed approach effectively suppresses geometric tail risks and significantly improves task success rates on a real robotic system.

Extensive experiments on a UR5 platform demonstrate that the proposed method reduces failure rates, decreases execution variance, and mitigates rare but catastrophic errors, particularly under near-field and off-axis configurations. The proposed mechanism is estimator-agnostic and can be readily integrated with both classical and learning-based pose estimation pipelines, providing a practical and general solution for improving execution-level reliability in vision-guided robotic systems.